%% file: main.tex
\documentclass[submission,copyright,creativecommons]{eptcs}
\providecommand{\event}{ESSLLI 2022} 

\usepackage{iftex}

\usepackage[utf8]{inputenc}
\usepackage{graphicx}
\usepackage[dvipsnames]{xcolor}
\usepackage{amsmath}
\usepackage{braket}
\usepackage{textcomp}
\usepackage{hyperref}
\usepackage{amssymb}
\usepackage{amsthm}
\usepackage{verbatim}
\usepackage{adjustbox}

\usepackage{tikz-cd}
\usepackage{macros/tikzfig}
\input{macros/gramdiag.tikzstyles}

\usepackage{caption}
\usepackage{subcaption}

\newtheorem{theorem}{Theorem}
\newtheorem{lemma}[theorem]{Lemma}

\title{Language-independence of DisCoCirc's Text Circuits:\\ English and Urdu}
\author{Muhammad Hamza Waseem
\institute{Clarendon Laboratory, Department of Physics\\ University of Oxford}
\institute{Compositional Intelligence Team\\
Quantinuum}
\email{hamza.waseem@physics.ox.ac.uk}
\and
Jonathon Liu
\institute{Compositional Intelligence Team\\
Quantinuum}
\email{jonathon.liu@cambridgequantum.com}
\and
Vincent Wang-Ma\'{s}cianica
\institute{Quantum Group, Department of Computer Science\\
University Oxford}
\institute{Compositional Intelligence Team\\
Quantinuum}
\email{vincent.wang@cambridgequantum.com}
\and
Bob Coecke
\institute{Compositional Intelligence Team\\
Quantinuum}
\email{bob.coecke@cambridgequantum.com}
}
\def\titlerunning{Language-independence of DisCoCirc's Text Circuits: English and Urdu}
\def\authorrunning{M.H. Waseem, J. Liu, V. Wang-Ma\'{s}cianica \& B. Coecke}

\begin{document}
\maketitle

\begin{abstract}
DisCoCirc \cite{coecke2021mathematics} is a newly proposed framework for representing the grammar and semantics of texts using compositional, generative circuits. While it constitutes a development of the Categorical Distributional Compositional (DisCoCat) framework, it exposes radically new features. In particular, \cite{wang2022} suggested that DisCoCirc goes some way toward eliminating grammatical differences between languages.

In this paper we provide a sketch that this is indeed the case for restricted fragments of English and Urdu.  We first develop DisCoCirc for a fragment of Urdu, as it was done for English in \cite{wang2022}.  There is a simple translation from English grammar to Urdu grammar, and vice versa. We then show that differences in grammatical structure between English and Urdu - primarily relating to the ordering of words and phrases - vanish when passing to DisCoCirc circuits.
\end{abstract}

\section{Introduction}

In \cite{coecke2021mathematics} Coecke introduced the Compositional Distributional Circuits (DisCoCirc) framework for modelling the structure of meaning in natural languages, building further on earlier work on the Categorical Distributional Compositional model for combining grammar and semantics \cite{CSC, FrobMeanI}. An important distinction between the two is that DisCoCirc is able to model not only the meaning of individual sentences, but also the interaction of sentences giving rise to the meaning of texts generally. The central idea is that the information associated with noun entities appearing in the text (encoded in circuit wires) are updated by sentences (modelled as gates) as the text progresses. 

DisCoCirc admits a two-dimensional string diagrammatic formalism, inspired by quantum circuits/networks, and therefore provides prospects for quantum-computational natural language processing for texts comprising multiple sentences or paragraphs \cite{QNLP-foundations}. It also has been applied to a number of problems including spatio-temporal models of language meaning \cite{TalkSpace}, logical and conversational negation in natural language \cite{rodatz2021conversational, shaikh2021composing}, and solving logical puzzles \cite{semspace_tiffany_2020, duneau2021parsing}. The relationship between DisCoCirc and discourse-representation theory \cite{kamp2013discourse} is also discussed in \cite{wang2022}. 

Recently, Wang-Ma\'{s}cianica, Liu and Coecke proposed a method for generating DisCoCirc diagrams for a significant fragment of English \cite{wang2022}. They started by creating a hybrid grammar for English text, incorporating `phrase structure', `pronominal links', `phrase regions', etc.
These hybrid grammar representations of text were then translated into DisCoCirc text circuits, via an intermediate structure called `text diagrams' that involving string diagrams.
The entire translation process preserves the compositionality and connectedness of text meaning. 

The same work describes how in the reverse direction, text circuits can be used as a generative grammar. For each freely generated text circuit, we can write some corresponding text. In this paper, we use `text' to refer not only to a string of words forming sentences, but to this string of words further endowed with a hybrid grammar structure. The text corresponding to a given text circuit is not unique owing to the loss of grammatical `bureaucracy' \cite{wang2022} in the passage from one-dimensional syntax to two-dimensional text circuits.\footnote{Some of this was already observed in \cite{GramEqs}.}
Here, grammatical bureaucracy is used in a broad sense to refer to all of the stylistic choices one must make when communicating some desired meaning in the form of a text.
That is, whenever two different texts communicate what is essentially the same meaning, we attribute the differences in the structure of these texts to grammatical bureaucracy.
Thus grammatical bureaucracy includes syntactic rules like those governing word order, but also choices like the use of pronouns, whether to use a single long sentence with multiple clausal constructions or multiple short sentences, etc.

It was in particular suggested that because of eliminating grammatical bureaucracy and stylistic choices embedded in a particular natural language, text circuits are to some degree language-independent. We take this suggestion seriously in this paper, and provide an outline of how DisCoCirc undoes the grammatical bureaucracy relating to word and phrase order for restricted fragments of English and Urdu. Our main argument is structured as follows.

\begin{itemize}
    \item In \cite{wang2022}, a hybrid grammar was developed for English. A surjection from the set of all English text generated with the English hybrid grammar to the set of all English text circuits was demonstrated:
    \[ \text{English text} \twoheadrightarrow \text{English text circuits} \]
    \item In a similar vein, in this paper, we describe how the hybrid grammar can be adapted for Urdu. We then provide rules for its translation into text diagrams and text circuits, which is essentially the same as in \cite{wang2022}. We show that this gives a surjective map from the set of all Urdu text generated with Urdu hybrid grammar to the set of all Urdu text circuits:
    \[ \text{Urdu text} \twoheadrightarrow \text{Urdu text circuits} \]
    \item For these restricted fragments, there is a clear isomorphism between the between the hybrid grammars for English and Urdu: 
    \[ \text{English hybrid grammar} \simeq \text{Urdu hybrid grammar} \] 
    \item Given the above correspondence, it turns out that text circuits for English and Urdu become the same, up to translating the labels on the gates. In other words, the following diagram commutes:
    \[
    \begin{tikzcd}
    \text{English text} \arrow[r, twoheadrightarrow] & \text{English Circuits} \arrow[d, leftrightarrow, "Dictionary\qquad\ \ "] \\ 
    \text{Urdu text} \arrow[u, leftrightarrow, "\begin{matrix}
    Grammar\  \&\
    Dictionary
    \end{matrix}"] \arrow[r, twoheadrightarrow] & \text{Urdu Circuits}
    \end{tikzcd}
    \]
    Now formally speaking, our texts consist of a hybrid grammar structure with labels (words) provided by a dictionary. If we restrict to just the grammatical structures and forget the language-specific word labels, the circuits for English and Urdu become literally the same. In this case, the above diagram reduces to:
    
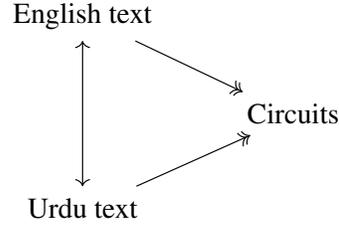
\[
    \begin{tikzcd}
    \text{English text} \arrow[dr, twoheadrightarrow] \arrow[dd, leftrightarrow] &  \\
    & \text{Circuits} \\
    \text{Urdu text} \arrow[ur, twoheadrightarrow] &
    \end{tikzcd}
    \]
\end{itemize}

The paper is organised as follows. The hybrid grammar, text diagrams and text circuits for English, along with an example, are reviewed in Section \ref{hyb_gram}. Hybrid grammar for Urdu is introduced in Section \ref{Urdu_gram}, followed by presentation of our sketch proof through an expository example in Section \ref{circ_Urdu}. Section \ref{conclusion} concludes the paper.

\section{Grammar, diagrams and circuits for English text}\label{hyb_gram}

In this section we quickly review the hybrid (generative) grammar for English developed in \cite{wang2022}.  First, simple sentences were modelled, containing verbs, adverbs, adjectives and adpositions. Then, pronominal links were introduced to account for recurring nouns and pronoun-referent pairs. Rewrite rules were introduced that allowed for the fusion of simple sentences into more complex ones, and the introduction of relative pronouns. Rules for modeling verbs with sentential complement and `phrase scope boundary' were introduced to accommodate compound sentences formed of components which are themselves sentences.

The hybrid grammar begins with a standard phrase structure grammar that generates our simple sentences, based on a string rewrite system with finitely many production rules of form $\alpha \mapsto \beta$.
These are valid transformations of strings of symbols represented by $\alpha$ and $\beta$. Individual symbols may be phrase components or entire words of the language we are modelling. In the latter case, the symbol is terminal, meaning no more rewrite rules can be further applied. 
In a grammar such as ours, a particular language comprises all the strings of terminal symbols that can be generated by applying finitely many production rules (associated with that language) to a start symbol, $\texttt{S}$. For example, using the rules:
\begin{equation*}
\begin{split}
    \texttt{S} &\mapsto \texttt{NP}_1 \cdot \texttt{TVP} \cdot \texttt{NP}_2 \\
    \texttt{NP}_1 &\mapsto \text{John} \\
    \texttt{TVP} &\mapsto \text{reads} \\
    \texttt{NP}_2 &\mapsto \text{books}
\end{split}    
\end{equation*}
where $\texttt{NP}$ and $\texttt{TVP}$ represent noun phrase and transitive verb phrase respectively; we can generate sentences like `John reads books': 
\begin{equation*}
\begin{split}
    \texttt{S} &\mapsto \texttt{NP}_1 \cdot \texttt{TVP} \cdot \texttt{NP}_2 \\
     &\mapsto \text{John} \cdot \texttt{TVP} \cdot \texttt{NP}_2 \\
     &\mapsto \text{John} \cdot \text{reads} \cdot \texttt{NP}_2 \\
     &\mapsto \text{John} \cdot \text{reads} \cdot \text{books} 
\end{split}    
\end{equation*}
Rewriting of strings can be represented as two-dimensional tree diagrams, read from top to bottom. Our example sentence can be represented as
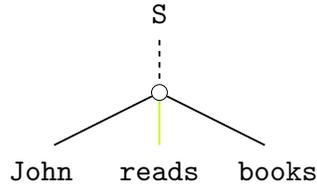
\[\scalebox{0.7}{\input{johnreadsbook.tikz}}\]

There may be multiple edges between parent and child nodes. Symbolic labels can be dropped for intermediate edges and, instead, a color coding can be used. In \cite{wang2022}, production rules and the corresponding planar trees were defined for simple sentences with verbs, adjectives, adverbs and adpositions, and compound sentences formed with pronominal links and phrase scope structures. 

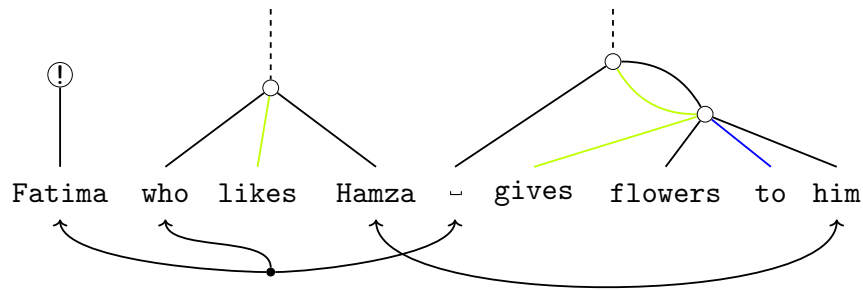
\[\scalebox{0.7}{\input{fatimawho.tikz}}\]

To do away with artefacts such as those handling pronominal links, \textit{text diagrams} are introduced. Based on string diagrams \cite{BaezLNP, SelingerSurvey, CKbook}, which is a structural framework for boxes with inputs and outputs, it allows parallel and sequential composition. $\texttt{S}$-type is replaced by $\texttt{NP}$-types, the number of which depends on the sentence. This modification allows composition of tree fragments (while respecting the grammatical types). Moving to text diagrams, pronominal links and phrase scope constructions become part of one unified mathematical framework.

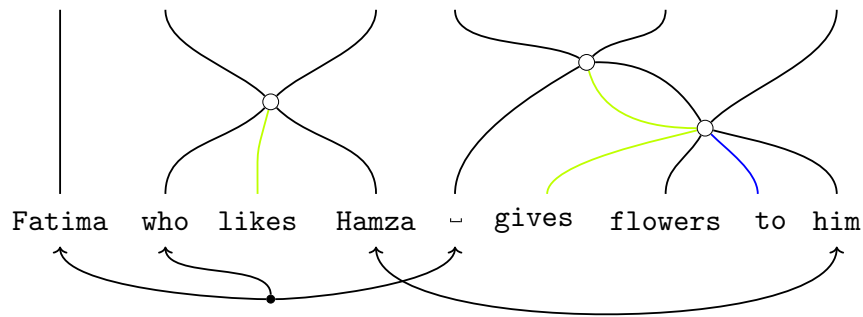
\[\scalebox{0.7}{\input{fatima2.tikz}}\]

Finally, rewrite rules are introduced to map fragments of text diagrams to \textit{text circuits}. Noun types are represented by wires. Adjectives and intransitive verbs are represented by single-input-single-output gates, and transitive verbs by double-input-double-output gates, acting on noun wires. Adverbs, adpositions and sentential complements, modifying verbs, are represented as boxes that contain the verb boxes/gates being modified. Conjunctions are taken to be boxes which contain two sentence circuits that are being connected.

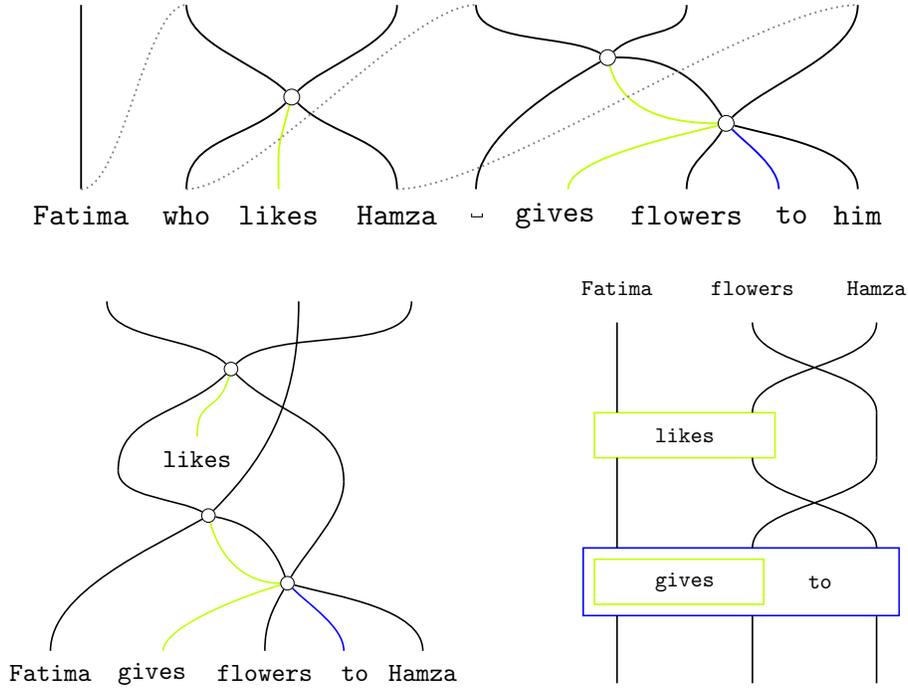
\[\scalebox{0.7}{\input{fatima3.tikz}}\]

\[\scalebox{0.6}{\input{fatima4.tikz}} \quad\quad\quad\quad \scalebox{0.6}{ \input{fatima5.tikz}}\]

It was shown in \cite{wang2022} that English text generated from the aforementioned hybrid grammar is surjective to text circuits. In this paper, we demonstrate a similar result for Urdu.

\section{A hybrid grammar for Urdu}\label{Urdu_gram}

Since a language is specified by set of production rules, different production rules lead to different languages. Translating to Urdu, `John reads books' can be transliterated\footnote{Urdu script is written from right to left, opposite to English. 
Throughout this paper, for ease of readability and linguistic analysis, we shall use English transliteration of Urdu text and hence use left-to-right script. 
} in English as
\begin{center}
\textit{John \ \ kitabein\ \ \ parhta hai} (Urdu) \\
\ John \ \ \ \   books \ \ \ \ \ \ \ reads  \ \ \ \ \ \ \ \ \ \ \ \ \ \ \ 
\end{center}

\noindent Using the production rules
\begin{equation*}
\begin{split}
    \texttt{S} &\mapsto \texttt{NP}_1 \cdot \texttt{NP}_2 \cdot \texttt{TVP} \\
    \texttt{NP}_1 &\mapsto \text{John} \\
    \texttt{NP}_2 &\mapsto \text{kitabein} \\
    \texttt{TVP} &\mapsto \text{parhta hai} \\
\end{split}    
\end{equation*}
we can generate the sentence under discussion
\begin{equation*}
\begin{split}
    \texttt{S} &\mapsto \texttt{NP}_1 \cdot \texttt{NP}_2 \cdot \texttt{TVP} \\
     &\mapsto \text{John} \cdot \texttt{NP}_2  \cdot \texttt{TVP} \\
     &\mapsto \text{John} \cdot \text{kitabein}  \cdot \texttt{TVP}  \\  
     &\mapsto \text{John} \cdot \text{kitabein}  \cdot \text{parhta  hai}
\end{split}    
\end{equation*}
the tree diagram of which is given by

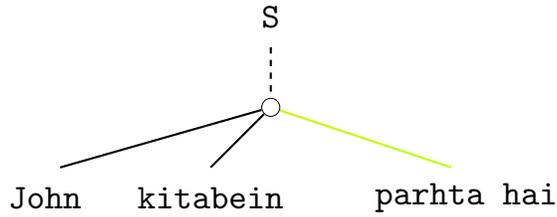
\[\scalebox{0.8}{\input{johnkitaabe.tikz}}\]

From this example already, we can spot an obvious difference between English and Urdu: the order of subject, verb and object. More particularly, the verb is usually placed at the end of the sentence in Urdu, unlike in English. In fact, as we shall see in this paper, this contrast plays a significant role in differentiating Urdu and English grammars.

We develop production rules and tree diagrams for the fragment of Urdu text (including verbs, adjectives, adverbs, adpositions, pronominal links, phrase scope) corresponding to that of English in \cite{wang2022}. Doing this, we realise that many of the rules and tree fragments are in fact the same. The ones that are different differ mainly in the relative placement of the verb. See Table \ref{tab:diff_gram}.

\section{Text diagrams and circuits for Urdu}\label{circ_Urdu}

\subsection{Main result}

Let $\texttt{E}$ denote the set of generators of the hybrid English grammar, summarised in Table \ref{tab:diff_gram}. 
Note we have restricted to a version of hybrid grammar generated by the explicitly context-free rules (i.e.,
we keep all generators except for the rule involving $\texttt{ADP}$ and $\texttt{TVP}$, and the rules that allow noun
wires to leave phrase scope).
Let $\texttt{T}_\texttt{E}$ denote the set of all English text constructed with grammar $\texttt{E}$, and let $\texttt{C}_\texttt{E}$ denote the set of all text circuits for English. Then, there exists a surjection $\texttt{T}_\texttt{E} \twoheadrightarrow \texttt{C}_\texttt{E}$ \cite{wang2022}.

We create a set of generators for Urdu grammar $\texttt{U}$ which closely correspond with the English generators $\texttt{E}$. Let $\texttt{T}_\texttt{U}$ denote the set of all Urdu text generated with $\texttt{U}$. Let $\texttt{C}_\texttt{U}$ denote the set of all text circuits for Urdu. Then,
\begin{itemize}
    \item there exists a surjection $\texttt{T}_\texttt{U}\twoheadrightarrow \texttt{C}_\texttt{U}$, and
    \item $\texttt{C}_\texttt{U}$ is isomorphic to $\texttt{C}_\texttt{E}$, i.e., $\texttt{C}_\texttt{U}
    \xrightarrow{\simeq}
    \texttt{C}_\texttt{E}$ (up to word translations at the gate level).
\end{itemize}

\subsection{Urdu text surjects onto circuits}

The method of turning hybrid grammar trees into text diagrams is the same in English and Urdu: each component of a hybrid grammar tree is modified so that the number of $\texttt{NP}$ wires for inputs and outputs is made equal, and sentence types $\texttt{S}$ are eliminated.

Table \ref{tab:diff_gram} illustrates the similarities and differences between Urdu and English grammar, as reflected in the hybrid grammar trees and text diagrams. The changes are as follows: 
\begin{itemize}
    \item Urdu has Subject-Object-Verb order, in contrast to Subject-Verb-Object as in English.
    \item The placement of adpositions differs from English; in Urdu, the verb simply comes last.    
    \item Sentential complements precede verbs in Urdu.
    \item In Urdu, the copula $\texttt{HAI}$ in the postpositional adjectival construction appears to the right of adjective. On the other hand, in English, the copula $\texttt{IS})$ appears to the left of the adjective. 
\end{itemize}

Everything else remains the same. For instance, the same reductions of text diagrams hold in Urdu as in English. Following is the reduction of a postpositional adjectival construction using a copula $\texttt{HAI} (\simeq \texttt{IS})$ to a prepositional adjective that does not require a copula.
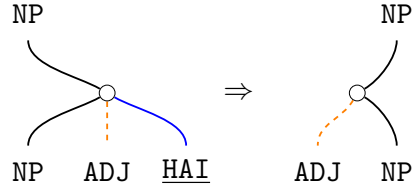
\[\scalebox{0.7}{\input{urdudiags/ADJ.tikz}}\]

The aforementioned modifications translate the formal claims of \cite{wang2022} for Urdu; any hybrid grammar text for Urdu surjects onto a text circuit.

\begin{table}
\centering
\scalebox{0.55}{
\begin{tabular}{|c|c|c|c|c|}
\hline
\text{Rule} & \text{English grammar} & \text{English diagram} & \text{Urdu grammar} & \text{Urdu diagram} \\ \hline
Intrans.Verb & \input{engdiags/IVcsg.tikz} & \input{engdiags/IVgraph.tikz} & \input{engdiags/IVcsg.tikz} & \input{engdiags/IVgraph.tikz} \\ \hline
\textcolor{red}{Trans.Verb} & \input{engdiags/TVcsg.tikz} & \input{engdiags/TVgraph.tikz} & \input{urdudiags/TVPgram.tikz} & \input{urdudiags/TVP.tikz} \\ \hline
Adjective(Pre.) & \input{engdiags/ADJcsg.tikz} & \input{engdiags/ADJgraph.tikz} & \input{engdiags/ADJcsg.tikz} & \input{engdiags/ADJgraph.tikz} \\ \hline
\textcolor{red}{Adjective(Post.)} &  \input{engdiags/ADJiscsg.tikz} &  \input{engdiags/ADJisgraph.tikz} & \input{urdudiags/ADJpostgrammar.tikz} & \input{urdudiags/ADJpostdiagram.tikz} \\ \hline
Adverb(IV) & \input{engdiags/ADVIVcsg.tikz} & \input{engdiags/ADVIVgraph.tikz} & \input{engdiags/ADVIVcsg.tikz} & \input{engdiags/ADVIVgraph.tikz}  \\ \hline
Adverb(TV) & \input{engdiags/ADVTVcsg.tikz} & \input{engdiags/ADVTVgraph.tikz} & \input{engdiags/ADVTVcsg.tikz} & \input{engdiags/ADVTVgraph.tikz} \\ \hline
\textcolor{red}{Adposition(IV)} & \input{engdiags/ADPIVcsg.tikz} & \input{engdiags/ADPiveng.tikz} & \input{urdudiags/ADPivgrammar.tikz}  & \input{urdudiags/ADPivdiagram.tikz} \\ \hline
\textcolor{red}{Sent.Comp.Verb} & \input{engdiags/SCVscopecsg.tikz} & \input{urdudiags/SCVeng.tikz} & \input{urdudiags/SCVgram.tikz} & \input{urdudiags/SCV.tikz}  \\ \hline
Conjunction & \input{engdiags/CNJscopecsg.tikz} & \input{urdudiags/CNJ.tikz}  & \input{engdiags/CNJscopecsg.tikz} & \input{urdudiags/CNJ.tikz}  \\ \hline
\end{tabular}
}
\caption{Generators for hybrid Urdu and English grammar and the corresponding diagrams. Note the different constructions for English and Urdu in the rules: Trans.Verb, Adjective(Post.), Adposition(IV) and Sent.Comp.Verb.}
\label{tab:diff_gram}
\end{table}

\subsection{English and Urdu give the same circuits}

In the case that we only consider context-free generators, the desired isomorphism between English and Urdu circuits essentially follows from the isomorphism between the trees generated by the English and Urdu hybrid grammar.

As a running example, we choose the English sentence `the young student who sees the honest teacher passionately teach smiles at him', which we translate into the Urdu sentence 
transliterated as:
\begin{center}
\textit{nojawan \ \ talib-e-ilm \ \ \  jo \ \ \ \ \ \ imandar \ \ \ \  ustad (ko) \ \ shauq se \ \ parhate huwe} (Urdu) \\
\ \ (the) young \ \ \  student \ \ \ \ who \ \ (the) honest \ \ \  teacher \ \   passionately \ \ \ \ \ \  teach \ \ \ \ \ \ \ \ \ \ \ \ \ \ \ \ \ \ \  \\
\ \\
\textit{dekhta hai \ \ us \ \ ki taraf \ \  muskurata hai} (Urdu) \\
\ \  \ sees  \ \ \ \ \ \ \   him \ \ \ \ \     at \ \ \ \ \ \ \ \  smiles \ \  \ \ \ \ \ \  \ \ \ \ \ \ \ \ \  \ \  
\end{center}

\begin{lemma}
Let $\texttt{T}_\texttt{E}$ denote the set of texts generated by the English production rules $\texttt{E}$, and $\texttt{T}_\texttt{U}$ denote the set of texts generated by the Urdu production rules $\texttt{U}$.
Viewing the syntax trees as rooted labelled trees (vertices are labels like $\texttt{NP}$ or $\underline{\texttt{HAI}}$, and the root is the initial sentence type $\texttt{S}$), there is an isomorphism $\texttt{T}_\texttt{E}\simeq \texttt{T}_\texttt{U}$ in the graph-theoretic sense.
\end{lemma}

Essentially, our Urdu generators $\texttt{U}$ correspond exactly to the English generators $\texttt{E}$, except some of them have the order of their outputs switched around.
So, given an English syntax tree generated by some sequence of applications of production rules in $\texttt{E}$, to translate this to an Urdu syntax tree we simply apply the corresponding rules in $\texttt{U}$ to the appropriate symbols. 
The exact proof of this is a simple induction.
Note that in order to `apply the right rule to the right symbol', we must track the identities of different symbols - e.g. distinguishing between different instances of $\texttt{NP}$ in our string. 
This tracking can be done by numbering the symbols with indices.
Note also the tracking of the identities of $\texttt{NP}$'s across wires is important later when we move to text diagrams, since we want to identify each `noun wire' in our diagram with a specific noun entity in our text. 


Now we have an isomorphism $\texttt{T}_\texttt{E}\simeq \texttt{T}_\texttt{U}$ between individual syntax trees.
Next we introduce pronominal links and the rewrite rules that allow us to adjoin pronominally linked trees into single sentences, thus attaining the full-blown `hybrid grammar' of \cite{wang2022}.
The isomorphism between trees lifts to a kind of isomorphism between these full hybrid grammar structures. 
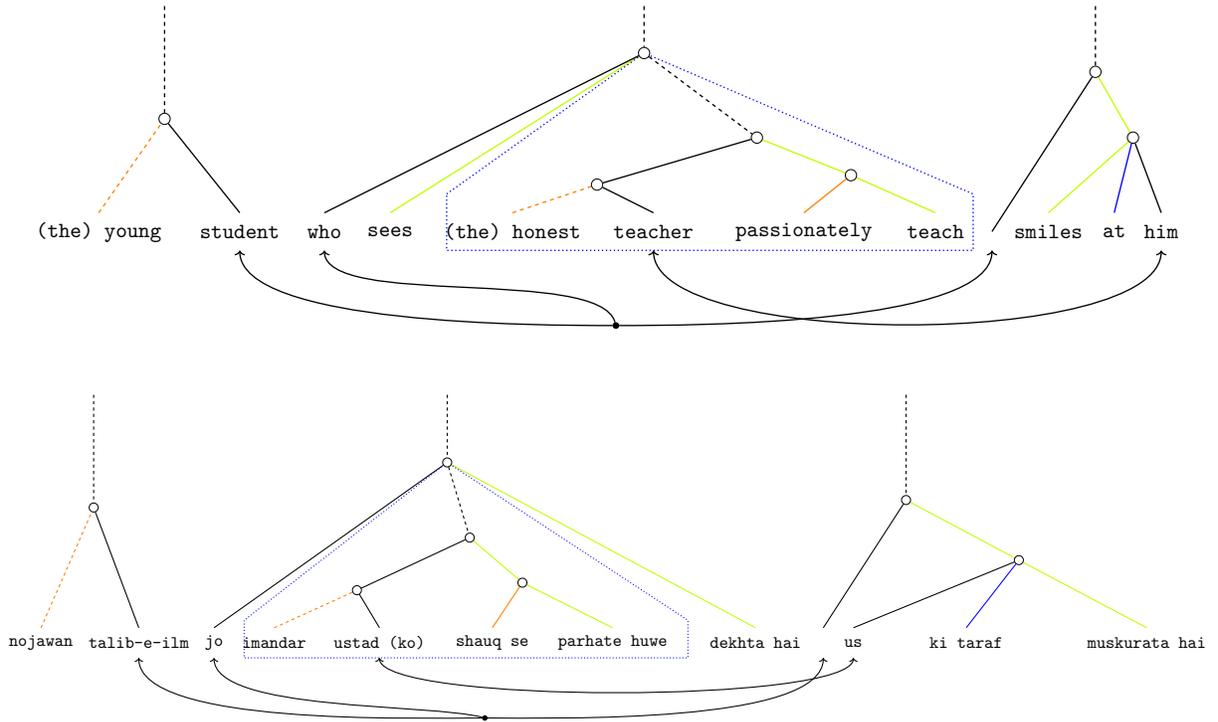
\[\scalebox{0.5}{\input{text2circENG_1.tikz}}\]
\[\scalebox{0.4}{\input{circ2textURD_6.tikz}}\]
Next we convert hybrid grammar to text diagrams.
We apply the rules in Table \ref{tab:diff_gram} for resolving the $\texttt{S}$ type into constituent $\texttt{N}$ wires and converting phrase scope into bubbles. Then we turn the pronominal links into dashed wires in preparation for composition.
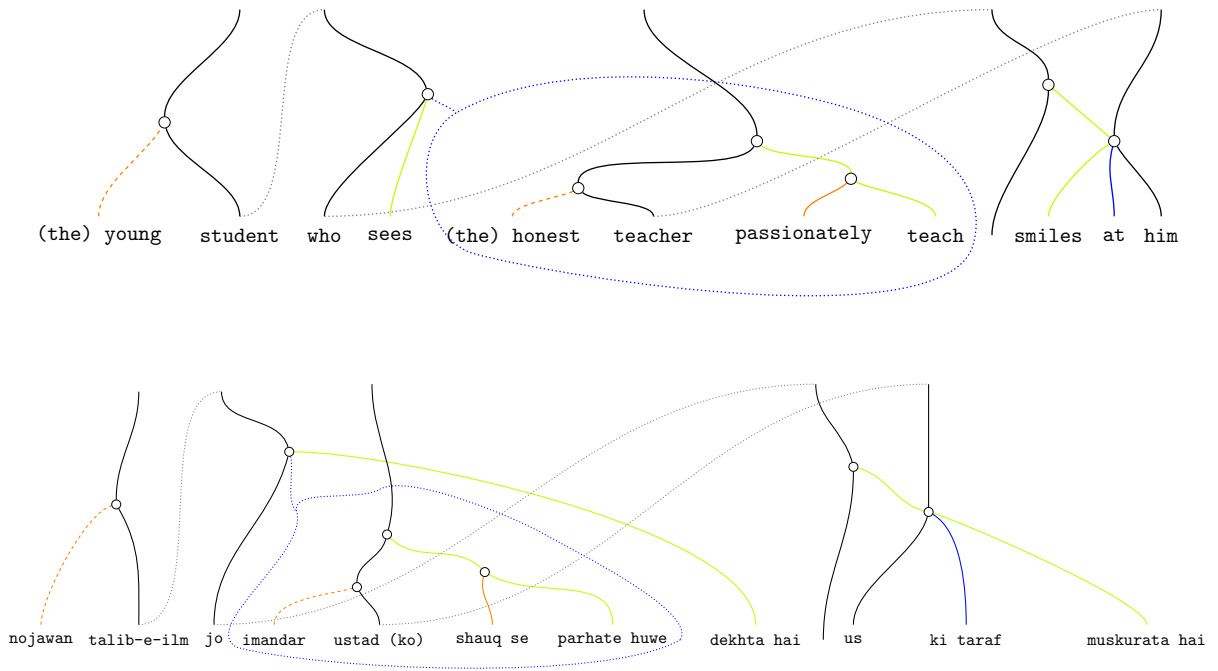
\[\scalebox{0.5}{\input{text2circENG_2.tikz}}\]
\[\scalebox{0.4}{\input{circ2textURD_5.tikz}}\]
After performing the composition, we recover text diagrams.
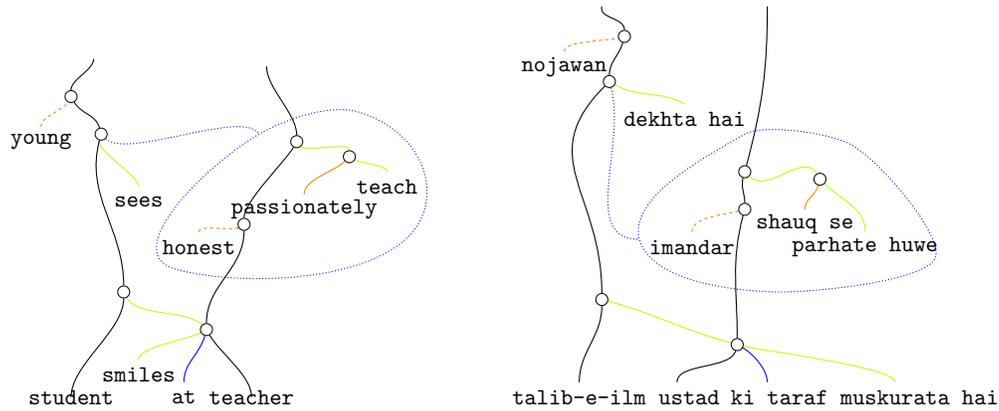
\[\scalebox{0.4}{\input{text2circENG_3.tikz} \quad\quad\quad\quad\quad\quad \input{circ2textURD_4.tikz}}\]
We see that the isomorphism of hybrid grammar structures simply lifts to an isomorphism of the structures at each intermediate step.
At the step where we reach the level of text diagrams, the isomorphism becomes an exact equality (modulo the different word labels, and allowing a certain degree of topological deformation as we usually do in string diagrams).

With the (topologically) identical structure of the English and Urdu diagrams, we simply apply the same conversion map from text diagrams to text circuits to obtain the same text circuit up to word-translations at the level of individual gates.
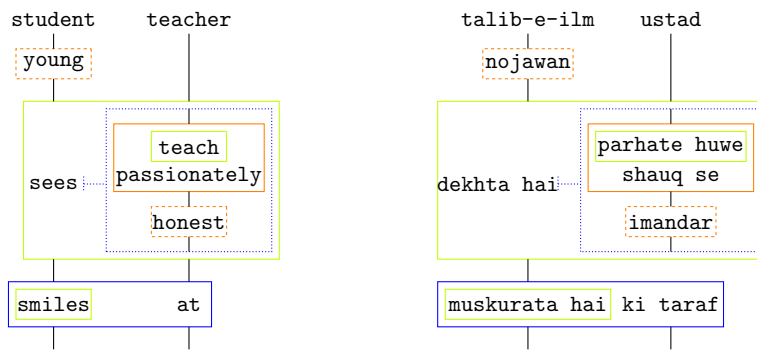
\[\scalebox{0.4}{\input{text2circENG_4.tikz} \quad\quad\quad\quad\quad\quad\quad\quad\quad\quad\quad\quad\quad \input{ENG2URDU.tikz}}\]
This process from text-to-circuits can be (nondeterministically) reversed, which make text circuits generative formalisms for English and Urdu text.




\section{Conclusion}\label{conclusion}






A major difference between grammars in different natural languages arises from different word orderings for, for example, subject, verb and object. For instance, in English the usual ordering is subject-verb-object, whereas in Urdu, it is subject-object-verb. These differences, in turn, exist because human verbal communication is restricted to one dimension and different cultures and demographics made different stylistic choices as languages evolved \cite{wang2022}. But there is no such restriction on machines. Two-dimensional grammars such as ours may be a suitable abstraction of text for computers (particularly quantum computers) and may prove advantageous for natural language processing tasks, such as machine translation. This paper moves a step forward in this direction.  
We emphasise that this paper represents a first step, and there is work to be done to expand the fragments of natural language that we can handle.
 
\section*{Acknowledgments}
Muhammad Hamza Waseem acknowledges the Rhodes Trust for funding his graduate studies at Oxford. 

\bibliographystyle{eptcs}
\bibliography{bibliography}

\end{document}

%% file: macros/gramdiag.tikzstyles
\tikzstyle{H}=[-, style=dashed]
\tikzstyle{K}=[-, line width=1pt]
\tikzstyle{Kv}=[-, line width=1pt, ->]
\tikzstyle{Kv<>}=[-,line width=1pt,{<->}]
\tikzstyle{gF}=[-, draw=none, fill={rgb,255: red,191; green,191; blue,191}]
\tikzstyle{KB}=[-, draw=blue, line width=1pt]
\tikzstyle{KO}=[-, draw={rgb,255: red,255; green,128; blue,0}, line width=1pt]
\tikzstyle{KL}=[-, draw={rgb,255: red,191; green,255; blue,0}, line width=1pt]
\tikzstyle{KHO}=[-, draw={rgb,255: red,255; green,128; blue,0}, style=dashed, line width=1pt]
\tikzstyle{KTG}=[-, draw={rgb,255: red,128; green,128; blue,128}, style=dotted, line width=1pt]
\tikzstyle{KTlG}=[-, draw={rgb,255: red,191; green,191; blue,191}, style=dotted, line width=1pt]
\tikzstyle{KBv}=[-, draw=blue, ->]
\tikzstyle{KOv}=[-, draw={rgb,255: red,255; green,128; blue,0}, line width=1pt, ->]
\tikzstyle{KLv}=[-, draw={rgb,255: red,191; green,255; blue,0}, ->]
\tikzstyle{T}=[-, style=dotted]
\tikzstyle{wF}=[-, fill=white, draw=none]
\tikzstyle{KH}=[-, style=dashed, line width=1pt]
\tikzstyle{Hv}=[->, style=dashed]
\tikzstyle{cv}=[-,right hook->]
\tikzstyle{vv}=[-,->>]
\tikzstyle{v}=[-,->]
\tikzstyle{<>}=[-,<->]
\tikzstyle{Hvv}=[-,->>,style=dashed]
\tikzstyle{Kvp}=[->]
\tikzstyle{KTB}=[-, draw=blue, style=dotted, line width=1pt]
\tikzstyle{KBgF}=[-, fill={rgb,255: red,191; green,191; blue,191}, draw=blue, line width=1 pt]
\tikzstyle{KBggF}=[-, fill={rgb,255: red,128; green,128; blue,128}, draw=blue, line width=1 pt]
\tikzstyle{b-wf}=[-, fill=white]
\tikzstyle{b-gf}=[-, fill={rgb,255: red,191; green,191; blue,191}, draw=black]
\tikzstyle{KHB}=[-, draw=blue, style=dashed, line width=1pt]

\tikzstyle{dot}=[inner sep=0mm,minimum width=2mm,minimum height=2mm,draw,shape=circle]
\tikzstyle{bigunit}=[dot,fill=white,text depth=-0.2mm]
\tikzstyle{smallblackdot}=[fill=black, inner sep=0mm,minimum width=1mm,minimum height=1mm,draw,shape=circle]
\tikzstyle{smallorangedot}=[fill={rgb,255: red,255; green,128; blue,0}, inner sep=0mm,minimum width=1mm,minimum height=1mm,draw,shape=circle]
\tikzstyle{smallbluedot}=[fill=blue, inner sep=0mm,minimum width=1mm,minimum height=1mm,draw,shape=circle]

%% file: main.bbl
\begin{thebibliography}{10}
\providecommand{\bibitemdeclare}[2]{}
\providecommand{\surnamestart}{}
\providecommand{\surnameend}{}
\providecommand{\urlprefix}{Available at }
\providecommand{\url}[1]{\texttt{#1}}
\providecommand{\href}[2]{\texttt{#2}}
\providecommand{\urlalt}[2]{\href{#1}{#2}}
\providecommand{\doi}[1]{doi:\urlalt{https://doi.org/#1}{#1}}
\providecommand{\eprint}[1]{arXiv:\urlalt{https://arxiv.org/abs/#1}{#1}}
\providecommand{\bibinfo}[2]{#2}

\bibitemdeclare{incollection}{BaezLNP}
\bibitem{BaezLNP}
\bibinfo{author}{J.~C. \surnamestart Baez\surnameend} \&
  \bibinfo{author}{M.~\surnamestart Stay\surnameend} (\bibinfo{year}{2011}):
  \emph{\bibinfo{title}{Physics, topology, logic and computation: a {R}osetta
  Stone}}.
\newblock In \bibinfo{editor}{B.~\surnamestart Coecke\surnameend}, editor:
  {\slshape \bibinfo{booktitle}{New Structures for Physics}},
  \bibinfo{series}{Lecture Notes in Physics}, \bibinfo{publisher}{Springer},
  pp. \bibinfo{pages}{95--172}, \doi{10.1007/978-3-642-12821-9\_2}.

\bibitemdeclare{misc}{QNLP-foundations}
\bibitem{QNLP-foundations}
\bibinfo{author}{B.~\surnamestart Coecke\surnameend},
  \bibinfo{author}{G.~\surnamestart de~Felice\surnameend},
  \bibinfo{author}{K.~\surnamestart Meichanetzidis\surnameend} \&
  \bibinfo{author}{A.~\surnamestart Toumi\surnameend} (\bibinfo{year}{2020}):
  \emph{\bibinfo{title}{Foundations for Near-Term Quantum Natural Language
  Processing}}.
\newblock \eprint{2012.03755}.

\bibitemdeclare{book}{CKbook}
\bibitem{CKbook}
\bibinfo{author}{B.~\surnamestart Coecke\surnameend} \&
  \bibinfo{author}{A.~\surnamestart Kissinger\surnameend}
  (\bibinfo{year}{2017}): \emph{\bibinfo{title}{Picturing Quantum Processes. A
  First Course in Quantum Theory and Diagrammatic Reasoning}}.
\newblock \bibinfo{publisher}{Cambridge University Press},
  \doi{10.1017/9781316219317}.

\bibitemdeclare{incollection}{CSC}
\bibitem{CSC}
\bibinfo{author}{B.~\surnamestart Coecke\surnameend},
  \bibinfo{author}{M.~\surnamestart Sadrzadeh\surnameend} \&
  \bibinfo{author}{S.~\surnamestart Clark\surnameend} (\bibinfo{year}{2010}):
  \emph{\bibinfo{title}{Mathematical foundations for a compositional
  distributional model of meaning}}.
\newblock In \bibinfo{editor}{J.~\surnamestart van Benthem\surnameend},
  \bibinfo{editor}{M.~\surnamestart Moortgat\surnameend} \&
  \bibinfo{editor}{W.~\surnamestart Buszkowski\surnameend}, editors: {\slshape
  \bibinfo{booktitle}{A Festschrift for Jim Lambek}}, {\slshape
  \bibinfo{series}{Linguistic Analysis}}~\bibinfo{volume}{36}, pp.
  \bibinfo{pages}{345--384}.
\newblock \eprint{1003.4394}.

\bibitemdeclare{article}{GramEqs}
\bibitem{GramEqs}
\bibinfo{author}{B.~\surnamestart Coecke\surnameend} \&
  \bibinfo{author}{V.~\surnamestart Wang\surnameend} (\bibinfo{year}{2021}):
  \emph{\bibinfo{title}{Grammar Equations}}.
\newblock \eprint{2106.07485}.

\bibitemdeclare{incollection}{coecke2021mathematics}
\bibitem{coecke2021mathematics}
\bibinfo{author}{Bob \surnamestart Coecke\surnameend} (\bibinfo{year}{2021}):
  \emph{\bibinfo{title}{The mathematics of text structure}}.
\newblock In: {\slshape \bibinfo{booktitle}{Joachim Lambek: The Interplay of
  Mathematics, Logic, and Linguistics}}, \bibinfo{publisher}{Springer}, pp.
  \bibinfo{pages}{181--217}, \doi{10.1007/978-3-030-66545-6\_6}.

\bibitemdeclare{misc}{semspace_tiffany_2020}
\bibitem{semspace_tiffany_2020}
\bibinfo{author}{T.~\surnamestart Duneau\surnameend} (\bibinfo{year}{2020}):
  \emph{\bibinfo{title}{{Solving} logical puzzles in {DisCoCirc}}}.
\newblock \urlprefix\url{https://www.youtube.com/watch?v=7Mri3OXzJq4}.

\bibitemdeclare{article}{duneau2021parsing}
\bibitem{duneau2021parsing}
\bibinfo{author}{T.~\surnamestart Duneau\surnameend} (\bibinfo{year}{2021}):
  \emph{\bibinfo{title}{Parsing Conjunctions in DisCoCirc}}.
\newblock {\slshape \bibinfo{journal}{SemSpace 2021}}, p.~\bibinfo{pages}{66}.

\bibitemdeclare{book}{kamp2013discourse}
\bibitem{kamp2013discourse}
\bibinfo{author}{H.~\surnamestart Kamp\surnameend} \&
  \bibinfo{author}{U.~\surnamestart Reyle\surnameend} (\bibinfo{year}{2013}):
  \emph{\bibinfo{title}{From discourse to logic: Introduction to modeltheoretic
  semantics of natural language, formal logic and discourse representation
  theory}}.
\newblock \bibinfo{volume}{42}, \bibinfo{publisher}{Springer Science \&
  Business Media}, \doi{10.1007/978-94-017-1616-1}.

\bibitemdeclare{article}{rodatz2021conversational}
\bibitem{rodatz2021conversational}
\bibinfo{author}{B.~\surnamestart Rodatz\surnameend}, \bibinfo{author}{R.~A.
  \surnamestart Shaikh\surnameend} \& \bibinfo{author}{L.~\surnamestart
  Yeh\surnameend} (\bibinfo{year}{2021}): \emph{\bibinfo{title}{Conversational
  Negation using Worldly Context in Compositional Distributional Semantics}}.
\newblock \eprint{2105.05748}.

\bibitemdeclare{article}{FrobMeanI}
\bibitem{FrobMeanI}
\bibinfo{author}{M.~\surnamestart Sadrzadeh\surnameend},
  \bibinfo{author}{S.~\surnamestart Clark\surnameend} \&
  \bibinfo{author}{B.~\surnamestart Coecke\surnameend} (\bibinfo{year}{2013}):
  \emph{\bibinfo{title}{The {F}robenius anatomy of word meanings {I}: subject
  and object relative pronouns}}.
\newblock {\slshape \bibinfo{journal}{Journal of Logic and Computation}}
  \bibinfo{volume}{23}, pp. \bibinfo{pages}{1293--1317},
  \doi{10.1093/logcom/ext044}.
\newblock \eprint{1404.5278}.

\bibitemdeclare{incollection}{SelingerSurvey}
\bibitem{SelingerSurvey}
\bibinfo{author}{P.~\surnamestart Selinger\surnameend} (\bibinfo{year}{2011}):
  \emph{\bibinfo{title}{A survey of graphical languages for monoidal
  categories}}.
\newblock In \bibinfo{editor}{B.~\surnamestart Coecke\surnameend}, editor:
  {\slshape \bibinfo{booktitle}{New Structures for Physics}},
  \bibinfo{series}{Lecture Notes in Physics},
  \bibinfo{publisher}{Springer-Verlag}, pp. \bibinfo{pages}{275--337},
  \doi{10.1007/978-3-642-12821-9\_4}.
\newblock \eprint{0908.3347}.

\bibitemdeclare{article}{shaikh2021composing}
\bibitem{shaikh2021composing}
\bibinfo{author}{R.~A. \surnamestart Shaikh\surnameend},
  \bibinfo{author}{L.~\surnamestart Yeh\surnameend},
  \bibinfo{author}{B.~\surnamestart Rodatz\surnameend} \&
  \bibinfo{author}{B.~\surnamestart Coecke\surnameend} (\bibinfo{year}{2021}):
  \emph{\bibinfo{title}{Composing Conversational Negation}}.
\newblock \eprint{2107.06820}.

\bibitemdeclare{article}{wang2022}
\bibitem{wang2022}
\bibinfo{author}{Vincent \surnamestart Wang\surnameend},
  \bibinfo{author}{Jonathon \surnamestart Liu\surnameend} \&
  \bibinfo{author}{Bob \surnamestart Coecke\surnameend} (\bibinfo{year}{2022}):
  \emph{\bibinfo{title}{Distilling Text into Circuits}}.
\newblock {\slshape \bibinfo{journal}{draft}}.

\bibitemdeclare{article}{TalkSpace}
\bibitem{TalkSpace}
\bibinfo{author}{V.~\surnamestart Wang-Mascianica\surnameend} \&
  \bibinfo{author}{B.~\surnamestart Coecke\surnameend} (\bibinfo{year}{2021}):
  \emph{\bibinfo{title}{Talking Space: inference from spatial linguistic
  meanings}}.
\newblock \eprint{2109.06554}.

\end{thebibliography}
